# Target Tracking In Real Time Surveillance Cameras and Videos

Nayyab Naseem

Department of Software Engineering

Fatima Jinnah Women University

abcxyz94786@gmail.com

Mehreen Sirshar

Department of Software Engineering

Fatima Jinnah Women University

msirshar@gmail.com

**Abstract:**

**Security concerns has been kept on increasing, so it is important for everyone to keep their property safe from thefts and destruction. So the need for surveillance techniques are also increasing. The system has been developed to detect the motion in a video. A system has been developed for real time applications by using the techniques of background subtraction and frame differencing. In this system, motion is detected from the webcam or from the real time video. Background subtraction and frames differencing method has been used to detect the moving target. In background subtraction method, current frame is subtracted from the referenced frame and then the threshold is applied. If the difference is greater than the threshold then it is considered as the pixel from the moving object, otherwise it is considered as background pixel. Similarly, two frames difference method takes difference between two continuous frames. Then that resultant difference frame is thresholded and the amount of difference pixels is calculated.**

## I.  Introduction:

Motion in images, videos carries important information about the change in the structure and dynamics of the scene. As it is natural phenomenon that motion always represents some activity both in the life of humans and animals. As animals always perceive the potential danger and the source of food that comes towards them by detecting its motion. Similarly, humans perceive the activity by sensing the physical movement. Tracking multiple targets such as humans, vehicles has a huge number of applications in different fields such as in video surveillance, airport security system and it has become an important area of research. Video cameras are the most extensively used sensors in the modern world ranges from selfies to surveillance.

Numerous algorithms such as background subtraction, frames differencing, and optical flow method have been developed to detect the moving target. Frame differencing method calculates the difference between two consecutive frames to detect the moving target. Optical flow method calculates the optical flow field of the image and then it performs clustering processing. Background subtraction method calculates the difference between the background image and the current image.

The recent works on target tracking has focused on the detection of current state of the target. These methods basically use current and the previous observations of the moving object to detect the current state. The method introduced in [1] proposes a tracking strategy that can track an object with the help of pre-recorded image database. The method introduced in [2] proposes a discrete continuous CRF approach for multi target tracking. The technique proposed in [5] selects a person in one view and recognize it in the other view. Method introduced in [6] proposes a local SVM approach for detecting human actions.

The aim of this research is to study the various methods used for the motion detection and tracking. The method that we propose for motion detection and tracking is based on background modeling and frame differencing. These algorithms are simple and easy to understand and relatively accurate.

## II. Methodology:

The main aim of this system is to detect motion in different scenes and to improve surveillance techniques.

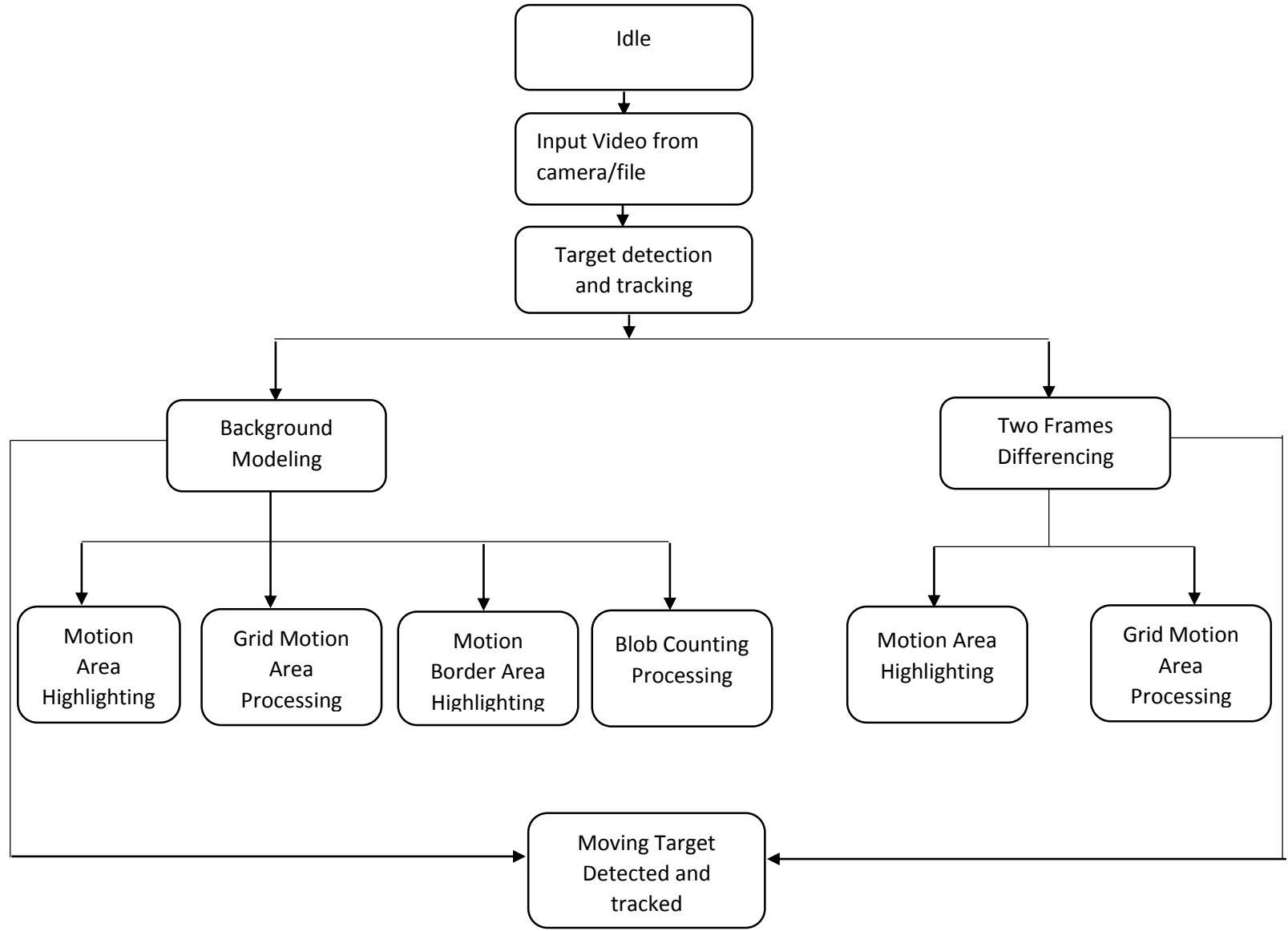

**Fig 1: Overview of the system**

Proposed system works on two methods:

- Two Frames differencing
- Background Modeling

First the system takes video as an input either through camera or from a file. Then the system applies two frame differencing and simple background modeling method on the video and tracks the target. The two frames difference algorithm works on difference of two continuous frames. Moving object is captured from a camera and difference is taken between the two continous frames. Then that resultant difference frame is thresholded and the amount of difference pixels is calculated. Using two frame differencing method, target can be tracked by means of motion area highlighting and grid motion area processing. Motion area highlighting method highlights the moving area with specified color. Similarly grid motion area processing method supports grid processing of the moving target. In this method, motion area has been divided into certain amount of cells and then motion level has been calculated for each cell.

Background modeling also known as background subtraction method works on the idea of taking the difference between the referenced frame that has been stored in the database and the current frame that has been grabbed from the camera. In background subtraction method, current frame is subtracted from the referenced frame and then the threshold is applied. If the difference is greater than the threshold then it is considered as the pixel from the moving object, otherwise it is considered as background pixel.

We represent background frame with the B (x, y) and foreground or current frame with the F (x, y). When we subtract background frame from the current frame we get:

$$R(x, y) = F(x, y) - B(x, y)$$

R (x, y) is the resultant frame.

The background modeling method says that if the pixel value is greater than the set threshold, then consider that pixel from the moving object, otherwise consider it as background pixel.

$$R(x, y) = \begin{cases} 1 & (F(x, y) - B(x, y)) > T \\ 0 & \text{otherwise} \end{cases}$$

Where T represents threshold value. The above mathematical expression indicates that after subtraction, if pixel value is greater than the set threshold value, then represent it with one otherwise with zero.

Using background modeling method, target can be tracked by means of motion area highlighting, grid motion area processing, motion border area highlighting and blob counting processing. Motion border area highlighting method is quite similar to motion area highlighting method except that it only highlights the border of the moving target. Blob counting method counts the neighboring pixels in the motion frame and represents them as a box of specified color.

## III. Experimental Result:

**Table 1: Comparison of Target Tracking Techniques**

| Research Paper | Precision | Accuracy |
|---|---|---|
| Ben Benfold et.al | 73.6% | 59.9% |
| Breitenstein et.al | 67.0% | 78.1% |
| Anton Milan et.al | 87.2% | 66.4% |
| Yi Yang et.al | - | 78.5% |
| Nayyab Naseem | 72.5% | 78% |

We have compared different target detection and tracking techniques and has deducted the result that background modeling and subtraction technique is a bettter approach for detecting target motion. The

## IV. Conclusion:

In this paper, a method based on background subtraction and frame differencing for detecting and tracking moving objects have been proposed. Using these methods, target detection and monitoring recent research is till focussing on how to make these techniques more better to get more precision and accurcay.

system has been developed which is able to detect moving targets and can effectively monitor them. This technique provides efficient approach for surveillance and it is aimed to be beneficial for security purposes. The proposed method is robust as it eliminated noise in detecting moving target

## V. References:

[1] Yiming Ye, John K. Tsotsos, Eric Harley, Karen Bennet, "Tracking a person with pre-recorded image database and a pan, tilt, and zoom camera", *Machine Vision and Applications (2000)*

[2] Ben Benfold, Ian Reid, "Stable Multi-Target Tracking in Real-Time Surveillance Video", *IEEE CVPR 2011 conference*

[3] Anton Milan, Konrad Schindler, Stefan Roth, "Detection- and Trajectory-Level Exclusion in Multiple Object Tracking", *IEEE Conf. on Computer Vision and Pattern Recognition (CVPR)*, Portland, OR, June 2013

[4] Piotr Dollar, Christian Wojek, Bernt Schiele, Pietro Perona, "Pedestrian Detection: An Evaluation of the State of the Art", *IEEE TRANSACTIONS ON PATTERN ANALYSIS AND MACHINE INTELLIGENCE*

[5] Martin Hirzer, Csaba Beleznai, Peter M. Roth, Horst Bischof, "Person Re-Identification by Descriptive and Discriminative Classification"

[6] Christian Schuldt, Ivan Laptev, Barbara Caputo, "Recognizing Human Actions: A Local SVM Approach", *Swedish Research Council*

[7] Sangmin Oh, Anthony Hoogs, Amitha Perera, Naresh Cuntoor, Chia-Chih Chen, Jong Taek Lee, Saurajit Mukherjee, J. K. Aggarwal, Hyungtae Lee, Larry Davis, Eran Swears, Xioyang Wang, Qiang Ji, Kishore Reddy, Mubarak Shah, Carl Vondrick, Hamed Pirsiavash, Deva Ramanan, Jenny Yuen, Antonio Torralba, Bi Song, Anesco Fong, Amit Roy-Chowdhury, Mita Desai, "A Large-scale Benchmark Dataset for Event Recognition in Surveillance Video"

[8] David A. Forsyth, Okan Arikan, Leslie Ikemoto, James O'Brien, Deva Ramanan5, "Computational Studies of Human Motion:Part 1, Tracking and Motion Synthesis", *Foundations and Trends in Computer Graphics and Vision*, Vol. 1, No 2/3 (2005) 77–254

[9] Deva Ramanan, "Part-based models for finding people and estimating their pose"

[10] IVAN LAPTEV, "Local Spatio-Temporal Image Features for Motion Interpretation", Stockholm, Sweden 2004

[11] Chaitanya Desai and Deva Ramanan, "Detecting Actions, Poses, and Objects with Relational Phraselets"

[12] Yi Yang, Deva Ramanan, "Articulated Human Detection with Flexible Mixtures-of-Parts"

[13] Rahul Garg, Deva Ramanan, "Where's Waldo: Matching People in Images of Crowds"

[14] Deva Ramanan, David Forsyth, Andrew Zisserman, "Tracking People and Recognizing Their Activities"

[15] Fabian Nater, "Abnormal Behavior Detection in Surveillance Videos"

[16] Ms Jyoti J. Jadhav, "Moving Object Detection and Tracking for Video Survelliance", Volume 2, Issue 4, June-July, 2014